\documentclass[letterpaper, 10 pt, conference]{ieeeconf}
\IEEEoverridecommandlockouts
\usepackage[english]{babel}
\usepackage[pdftex]{graphicx}
\usepackage{adjustbox}
\graphicspath{{./figures/}}
\DeclareGraphicsExtensions{.pdf,.jpg,.jpeg,.png}
\usepackage[cmex10]{amsmath}
\usepackage{bm}
\interdisplaylinepenalty=2500{}
\usepackage{physics}
\usepackage{booktabs,multirow}
\usepackage[caption=false,font=footnotesize]{subfig}

\usepackage{tikz,pgfplots}
\pgfplotsset{compat=1.3}
\usepackage{url}
\newcommand{\fref}[1]{Fig.~\ref{#1}}

\newcommand{\tref}[1]{Table~\ref{#1}}
\newcommand{\sref}[1]{Section~\ref{#1}}

\usepackage[acronym,shortcuts]{glossaries}
\glsdisablehyper
\newacronym{lrf}{LRF}{Laser Range Finder}
\newacronym{svd}{SVD}{Singular Value Decomposition}
\newacronym[longplural=Degrees of Freedom]{dof}{DoF}{Degree of Freedom}
\glsunset{dof}

\title{\LARGE
  \textbf{SCALAR - Simultaneous Calibration of 2D Laser And Robot's Kinematic Parameters Using Three Planar Constraints}}

\author{Teguh Santoso Lembono, Francisco Su\'{a}rez-Ruiz, and Quang-Cuong Pham%
  \thanks{The authors are with the School of Mechanical and Aerospace
          Engineering, Nanyang Technological University, Singapore.}}

\begin{document}
\maketitle
\thispagestyle{empty}
\pagestyle{empty}

\begin{abstract}
Industrial robots are increasingly used in various applications where the robot accuracy becomes very important, hence calibrations of the robot's kinematic parameters and the measurement system's extrinsic parameters are required. However, the existing calibration approaches are either too cumbersome or require another expensive external measurement system such as laser tracker or measurement spinarm. In this paper, we propose SCALAR, a calibration method to simultaneously improve the kinematic parameters of a 6-\ac{dof} robot and the extrinsic parameters of a 2D \ac{lrf} which is attached to the robot. Three flat planes are placed around the robot, and for each plane the robot moves to several poses such that the \ac{lrf}'s ray intersect the respective plane. Geometric planar constraints are then used to optimize the calibration parameters using Levenberg-Marquardt nonlinear optimization algorithm. We demonstrate through simulations that SCALAR can reduce the average position and orientation errors of the robot system from 14.6mm and 4.05$^o$ to 0.09mm and 0.02$^o$.  
\end{abstract}


\section{Introduction}
\label{sec:introduction}

In traditional robotics applications such as pick and place, spray-painting and spot-welding, the robots either do not need very high accuracy or they are programmed by teaching, where the \textbf{repeatability} of the robot is more important than the \textbf{accuracy}. Repeatability refers to the robot's capability to return precisely to the same location as previously taught, whereas accuracy refers to the robot's capability to precisely reach a pose computed based on the robot's kinematic model. 

However, there are many applications where the accuracy of the robot becomes very crucial, given that the robot has to adapt to each task automatically with a great precision. Consider, for example, a robot drilling task in \cite{Suarez-Ruiz2018} where the robot is supposed to drill several holes at precisely-defined locations on a workpiece. The workpiece can be different for each task, and the placement within the workspace may not be precisely known. Programming by teaching in this case requires manual re-programming for each workpiece which is very inefficient. To program the robot automatically for such task, the robot has to do a few things accurately: the robot has to scan the workpiece, determine the location of the holes, and finally move to that location accurately. The accuracy of such a robotic system depends on at least two things: The accuracy of the robot and the accuracy of the measurement system. 
\begin{figure}[t]
  \centering
  \vspace*{2mm}
  \subfloat[]{\includegraphics[height=60mm]{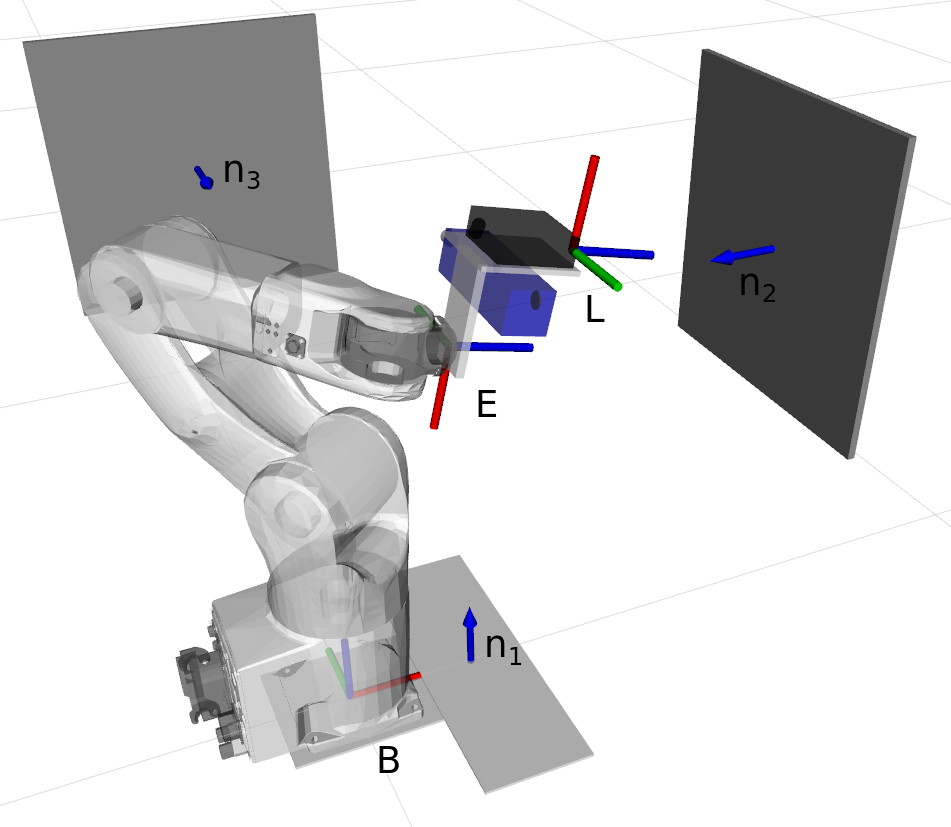}}
  \caption{Calibration Setup}
  \label{fig:robot_setup}
\end{figure}

The accuracy of the robot is determined by how closely the kinematic parameters of the robot's model resemble the actual kinematic parameters of the physical robot. This is affected by the manufacturing process, the assembly process, and the wear and tear during the operation of the robot. \textbf{Robot kinematic calibration} is usually conducted to achieve a better accuracy, either by using an external measurement system (such as motion capture system or coordinate measuring machines) or by constraining the motion of the end-effector.

The accuracy of the measurement system can be divided into two parts: the accuracy of the measurement device itself and the accuracy of the relative pose between the robot frame and the measurement frame. The accuracy of the measurement device depends on the type of device that is used and its specification. For example, a camera system is generallly less precise as compared to a laser system, although a camera can give more information. The second part of the accuracy comes from the fact that the data from a measurement system is always obtained in the measurement device coordinate frame, and it needs to be transformed to the robot coordinate system. Hence, the relative pose between the robot coordinate frame and the measurement device frame needs to be obtained. This relative pose is often called the \textbf{extrinsic parameters} of the measurement device, and the method to estimate the extrinsic parameters is called \textbf{extrinsic calibration}.

\renewcommand{\arraystretch}{1.3}
\begin{table*}[htp]
\caption{Examples of Unconstrained Calibration}
\label{tab:unconstrained_calib}
\centering
\begin{tabular}{c c c c c}
\toprule
\textbf{Researchers} &  \textbf{Robot} & \textbf{Measurement Device} &  \textbf{Initial Accuracy[mm]}  & \textbf{Final Accuracy[mm]}\\
\midrule
Ginani and Mota \cite{Ginani2011} & ABB IRB 2000 & ROMER Measurement Arm & 2.20 & 1.40 \\
Ye et al. \cite{Ye2006} & ABB IRB 2400/L & Faro Laser Tracker & 1.764 & 0.640 \\
Nubiola and Bonev \cite{Nubiola2013} & ABB IRB 1600-6/1.45 & Faro Laser Tracker ION & 2.158 & 0.696 \\ 
Newman et al. \cite{Newman2000} & Motoman P-8 & SMX Laser Tracker & 3.595 & 2.524\\
\bottomrule
\end{tabular}
\end{table*}

In this paper we present SCALAR, a calibration method to simultaneously improve the accuracy of the robot and the measurement system. SCALAR calibrates simultaneously the kinematic parameters of the robot and the extrinsic parameters of a 2D \acl{lrf} (\ac{lrf}) using only the information provided by the \ac{lrf} attached to the robot end-effector. An \ac{lrf} is chosen because it gives very accurate measurement data both for the calibration and for the subsequent tasks (such as drilling).  

The overall calibration procedure is as follows:
\begin{enumerate}
\item The \ac{lrf} is attached to the robot and three perpendicular planes are placed around the robot as shown in \fref{fig:robot_setup}. Only the rough estimate of the position and orientation of the planes are necessary to be known, so the setup can be easily done.
\item For each plane, the robot is moved to several poses such that the \ac{lrf}'s 2D ray is projected onto the plane. The data from the \ac{lrf} as well as the robot's joint angles information are collected.
\item An optimization algorithm is used to find the optimal robot's kinematic parameters together with the \ac{lrf} extrinsic parameters, using the geometric constraint that the projected \ac{lrf} data should be located on the three planes. 
\end{enumerate}

The remainder of the paper is as follows. In \sref{sec:related} we discuss the existing approaches to the calibration problem both for the robot's kinematic parameters and the \ac{lrf}'s extrinsic parameters, and how SCALAR differs from the other approaches. In \sref{sec:method}, SCALAR is explained in detail. A simulation study is presented in \sref{sec:simulation} to verify SCALAR, and finally we conclude with a few remarks in \sref{sec:conclusions}.

\section{Related works}
\label{sec:related}
\subsection{Calibration of robot's kinematic parameters}
\label{sec:kine_calib}

Robot kinematic calibration has been researched for a long time; some of the earliest works began in 1980s. The calibration procedures can be categorized into unconstrained and constrained calibration. In unconstrained calibration, the robot moves its end-effector to several poses while an external measurement system measures the pose. The measured pose is then compared to the one computed from the kinematic model, and the model is updated to minimize the difference between the predicted pose and the measured pose. In constrained calibration, some constraints are applied to the motion of the end-effector, and the constraints yield several calibration equations by which the robot's kinematic parameters are calibrated.

Examples of unconstrained calibration works can be seen in \tref{tab:unconstrained_calib}. The issues with such calibration method are the difficulty in setting up the calibration setup and the expensive cost of the external measurement system. For example, the cost of a laser tracker is more than $\$100,000$ USD \cite{Nubiola2013}. Therefore, many researchers try to find calibration methods which only rely on the internal sensors of the robot by constraining the motion of the end-effector such as in constrained calibration. 

In \cite{Ikits1997}, Ikits and Hollerbach propose a kinematic calibration method using a planar constraint. A touch probe attached to the robot's flange is moved to touch random points on a plane. When the touch probe is in contact with the plane, the joint angles are recorded. The kinematic parameters of the robot model are then updated to satisfy the planar constraint. While the approach is promising, they also report that some of the parameters are hardly observable when the measurements are noisy or when the model is not complete. The unobservability of the parameters means that some of the parameters cannot be obtained accurately from the calibration procedure.

In \cite{Zhuang1999}, Zhuang et al. investigate robot calibration with planar constraints, in particular the observability conditions of the robot's kinematic parameters. They prove that a single-plane constraint is insufficient for calibrating a robot, and a minimum of three planar constraints are necessary. Using three planar constraints, the constrained system is proved to be equivalent to an unconstrained point-measurement system under three conditions: a) All three planes are mutually non-parallel, b) the identification Jacobian of the unconstrained system is nonsingular, and c) the measured points from each individual plane do not lie on a line on that plane. They verify the theory by doing a simulation on a PUMA560 robot. 

In \cite{Joubair2015}, Joubair and Bonev calibrated both the kinematic and non-kinematic (stiffness) parameters of a FANUC LR Mate 200iC industrial robot by using planar constraints, in the form of a high precision 9-inches granite cube. The robot is equipped with an MP250 Renishaw touch probe, which is then moved to touch four planes of the granite cube. The granite cube's face is flat to within 0.002mm. They improved the maximum plane error from 3.740mm to 0.083mm. 

\subsection{Calibration of extrinsic 2D \ac{lrf} parameters}
\label{sec:laser_calib}
Extrinsic calibration of an \ac{lrf} consists of finding the correct homogeneous transformation from the robot coordinate frame to the laser coordinate frame. Most of the works on extrinsic calibration of an \ac{lrf} involves a camera, since both sensors are often used together in many applications. The works in this field are largely based on Zhang and Pless' work \cite{Zhang2004}. They propose a method to calibrate both a camera and an \ac{lrf} using a planar checkerboard pattern. First, the camera is calibrated by a standard hand-eye calibration  \cite{BouguetJ.Y.2003} using a checkerboard pattern. The calibrated camera is then used to calculate the pose of the pattern. Next, the robot is moved to several poses with the \ac{lrf} pointing to the pattern. By using the geometric constraints that all the data points from the \ac{lrf} should fall on the pattern plane, the extrinsic parameters of the \ac{lrf} can be obtained. Finally, the same constraints are used to optimize both the intrinsic and extrinsic parameters of the camera and the extrinsic parameters of the \ac{lrf}. The nonlinear optimization problem is solved by using Levenberg-Marquardt optimization algorithm.

Unnikrishnan and Hebert \cite{Unnikrishnan2005} use the same setup as \cite{Zhang2004}, although they do not optimize the camera parameter simultaneously due to the nonlinearity of the resulting cost function. 
Li et al. \cite{Li2007} use a specially designed checkerboard to calibrate the extrinsic parameters between a camera and an \ac{lrf}, and claim that the result is better than \cite{Zhang2004}. Vasconcelos et al. \cite{Vasconcelos2012} develop a minimal closed-form solution for the extrinsic calibration of a camera and an \ac{lrf}, based on the work in \cite{Zhang2004}. 

\subsection{Novelty of the proposed method}
\label{sec:novelty}
SCALAR can be seen as a combination of the algorithm for extrinsic calibration of an \ac{lrf} \cite{Zhang2004} and the algorithm for calibration of robot's kinematic parameters using three planar constraints \cite{Joubair2015}. It has the following advantages as compared to the other calibration approaches:
\begin{enumerate}
\item SCALAR simultaneously calibrates both the \ac{lrf} extrinsic parameters and the robot's kinematic parameters. Given that calibration process is often cumbersome, this saves a lot of time and effort. Moreover, the errors in the robot's kinematic parameters affect the extrinsic calibration accuracy, and vice versa. Hence, calibrating both parameters simultaneously results in better accuracy. 
\item SCALAR does not need an additional camera to calibrate the \ac{lrf}, unlike \cite{Zhang2004}.
\item SCALAR does not need another expensive external measurement system. The measurement is done using the \ac{lrf} that will also be used in the subsequent robot task, hence it does not incur additional cost. Moveover, an \ac{lrf} can achieve very high accuracy at much lower cost (more than ten times cheaper) as compared to the commonly used measurement systems such as Vicon or Faro Laser Tracker. 
\item SCALAR does not need a precisely manufactured calibration object such as the granite cube in \cite{Joubair2015}, which requires the planes' position and orientation to be known accurately. SCALAR only requires
three flat surfaces which are oriented roughly perpendicular to each other and the rough estimate of their locations. This also means that the calibration setup can be done easily.
\item The calibration poses can be distributed throughout the whole workspace, instead of being confined only in a local region such as in \cite{Joubair2015}. 
\end{enumerate}


\section{Method}
\label{sec:method}

The calibration setup is depicted in \fref{fig:robot_setup}, where three roughly perpendicular planes ($k=1,2,3$) are placed around the robot. An \ac{lrf} is attached to the robot flange. For each plane, the robot is moved to $N$ poses such that the \ac{lrf}'s ray is directed to the respective plane. One data set from the \ac{lrf} consists of hundreds of data points, so $M$ data points are selected from the \ac{lrf} data for each pose, and the robot's joint angles are recorded. 

This section describes the detail on how to calibrate both the robot's kinematic parameters and the \ac{lrf}'s extrinsic parameters. First, the initial estimate of the LRF's extrinsic parameters is obtained using the linear least-squares method with the data from one of the planes. This is based on the algorithm in \cite{Zhang2004}, although presented differently for better clarity. After that, the robot's kinematic parameters and the LRF's extrinsic parameters are optimized simultaneously to satisfy the three planar constraints using Levenberg-Marquardt nonlinear optimization method. Finally, we explain how \ac{svd} can be used to analyse which calibration parameters are identifiable, and the steps to handle the unidentifiable parameters are then presented. 
\subsection{Initial Estimate of the \ac{lrf} Extrinsic Parameters}
\label{sec:first_step}
To obtain an initial estimate of the \ac{lrf} extrinsic parameters, only the data from one plane is necessary. Arbitrarily, the bottom plane $P_1$ is chosen. The extrinsic parameters of the \ac{lrf} ${^E}\vb*{T}_L$, i.e. the homogeneous transformation from the robot flange coordinate frame to the \ac{lrf} coordinate frame, is estimated by the following calculations.

Let the subscript/superscript $B$, $E$, and $L$ denote the coordinate frame of the robot base, the robot flange, and the \ac{lrf}, while the subscript $i$, $j$, and $k$ refer to the \ac{lrf} data point index, the robot pose index, and the plane index respectively. Let $\vb*{p}_{ji}$ be one of the data points from the \ac{lrf} which lies on the $P_1$, $\vb*{n}_1$ be the normal unit vector of $P_1$, and $l_1$ be the perpendicular distance from the origin of the robot coordinate system to $P_1$.  Since $\vb*{p}_{ji}$ is located on the $P_1$, it has to satisfy the following constraint, 
  \begin{equation}
  \label{eq:1}
  {^B}\vb*{n}_1 ^T \cdot\; {^B}\vb*{p}_{ji}  - {^B}l_1 = 0 \; .
   \end{equation}
${^B}\vb*{p}_{ji}$ depends on the robot pose ${^B}\vb*{T}_{E,j}$ at pose index $j$ and the \ac{lrf} extrinsic parameter ${^E}\vb*{T}_L$, so \eqref{eq:1}  can be expanded,
  \begin{equation}
  {{^B}\vb*{n}_1 ^{T}} \cdot\; {^B}\vb*{T}_{E,j} \cdot \; {^E}\vb*{T}_L \cdot \; {^L}\vb*{p}_{ji}  - {^B}l_1 = 0 \; .
  \end{equation}
${^B}\vb*{n}_1$ and $^{B}l_1$ are known approximately ($[0 \; 0\; 1\;0]$ and $0.0$), ${^B}\vb*{T}_{E,j}$ can be computed from the robot's joint angles at pose index $j$, and ${^L}\vb*{p}_{ji}$ is obtained from the laser. Let 
\begin{equation}
{\vb*{n}_j^{'}}^{T} = {^B}\vb*{n}_1 ^T \cdot {^B}\vb*{T}_{E,j} = 
\left[n^{'}_{j,1} \quad n^{'}_{j,2} \quad n^{'}_{j,3}  \quad n^{'}_{j,4} \right] \; , 
\end{equation}
then  
  \begin{equation}
  \label{eq:3}
  {\vb*{n}_j^{'}} ^T \cdot {^E}\vb*{T}_L \cdot {^L}\vb*{p}_{ji} - {^B}l_1 = 0 \; .
  \end{equation}
The only unknown in \eqref{eq:3} is ${^E}\vb*{T}_L$ which has 12 elements $r_{uv}$, where u and v denote the column and the row index of the matrix. Note that the fourth row of ${^E}\vb*{T}_L$ only consists of 0 and 1. 
Without loss of generality, let's assume that the data points from the \ac{lrf} lie on the XZ planes of the laser frame $L$, so ${^L}\vb*{p}_{ji} = \left[{^L}p_{i,x} \quad 0 \quad {^L}p_{i,z}\quad 1\right]$. If we expand \eqref{eq:3} and rearrange such that the components of ${^E}\vb*{T}_L$ are stacked together as a vector $\vb*{\Phi}_L$, we have
\begin{equation}
\label{eq:4}
  {\vb*{x}_{ji}}^T \cdot \vb*{\Phi}_L = {^B}l_1 -  n^{'}_{j,4} \; , 
\end{equation}
where 
\begin{multline}
  \vb*{x}_{ji} = \left[{^L}p_{i,x}\;n^{'}_{j,1} \quad {^L}p_{i,x}\;n^{'}_{j,2}\quad {^L}p_{i,x}\;n^{'}_{j,3}\quad  {^L}p_{i,z}\;n^{'}_{j,1}\right. \\ 
\left. {^L}p_{i,z}\;n^{'}_{j,2}\quad {^L}p_{i,z}\;n^{'}_{j,3} \quad n^{'}_{j,1} \quad n^{'}_{j,2} \quad n^{'}_{j,3} \right]^T \; , 
\end{multline}
and
\begin{multline}
  \vb*{\Phi}_L= \left[r_{11} \quad r_{21} \quad r_{31} \quad r_{13} \quad r_{23} \quad r_{33} \quad r_{14} \quad r_{24}  \quad r_{34} \right] ^T \; .
\end{multline}
For each data point $i$, we obtain such equation as in \eqref{eq:4}. With $M$ data points per pose and a total of $N$ robot poses, there are $NM$ such equations. The equations can be stacked together to form the following matrix equation,
\begin{equation}
\label{eq:7}
  \vb*{X}   \vb*{\Phi}_L= \vb*{D} \; ,
\end{equation}
where 
\begin{equation}
\vb*{X} =\begin{bmatrix}
{\vb*{x}_{11}} \; \cdots \;{\vb*{x}_{1M}}\quad  {\vb*{x}_{21}} \; \cdots \; {\vb*{x}_{2M}}\;  \cdots \; {\vb*{x}_{NM}} 
\end{bmatrix}^T 
\end{equation}
and 
\begin{multline}
\vb*{D} =\left[
{^B}l_1 -  n^{'}_{1,4} \quad
\cdots \quad
{^B}l_1 -  n^{'}_{1,4} \quad
{^B}l_1 -  n^{'}_{2,4} \quad
\cdots  \quad
\right. \\
\left.
{^B}l_1 -  n^{'}_{2,4} \quad
\cdots \quad
{^B}l_1 -  n^{'}_{N,4} \quad  
\right]^T \quad .
\end{multline}
Equation \eqref{eq:7} can be solved by a linear least-square procedure to obtain $\vb*{\Phi}_L$. ${^E}\vb*{T}_L$ can then be computed from $\vb*{\Phi}_L$ as follows:
\begin{enumerate}
\item The parameters $[r_{11} \quad r_{21} \quad r_{31}]^T$ and $[r_{13} \quad r_{23} \quad r_{33}]^T$ are required to be unit vectors, so they have to be normalized. They consistute the first and the third column of the matrix ${^E}T_L$.
\item The parameters $[r_{14} \quad r_{24} \quad r_{34}]^T$ constitute the position component of the matrix ${^E}T_L$ (the 4\textit{th} column).
\item The parameters $[r_{12} \quad r_{22} \quad r_{32}]^T$ can be calculated as the cross product of  $[r_{13} \quad r_{23} \quad r_{33}]^T$ and $[r_{11} \quad r_{21} \quad r_{31}]^T$ .
\end{enumerate}

${^E}\vb*{T}_L$ has 12 parameters, but only 6 parameters are independent. To reduce the redundancy in the subsequent steps, the rotation part of ${^E}\vb*{T}_L$ is represented by the axis-angle representation $[r_x \quad r_y \quad r_z \quad r_{\theta}]$, while the position part is represented by  $[p_x \quad p_y\quad p_z]$.

\subsection{Optimizing both the \ac{lrf} Extrinsic Parameters and Robot's Kinematic Parameters}
\label{sec:second_step}
In the second step, the data from all the three planes are used to optimize the extrinsic parameters of the \ac{lrf}, the robot's kinematic parameters and the plane parameters. The objective function is described as follows:

\begin{equation}
 f (\vb*{\Phi}) =  \sum_{k=1}^{3} \sum_{j=1}^{N} \sum_{i=1}^{M} ({{^B}\vb*{n}_k}^T \cdot {^B}\vb*{p}_{ji} - {^B}l_k)^2
\end{equation}

The parameters $\vb*{\Phi}$ consist of the following:
\begin{enumerate}
\item Robot's kinematic parameters. We use  \textbf{DH} parameters \cite{Craig1986} $[a_i \;, \alpha_i \;,\theta_i \;,d_i], i=1, 2, \cdots ,6$ to represent the robot's kinematics, so there are 24 \textbf{DH} parameters for a 6-\ac{dof} robot arm. 
\item \ac{lrf}'s extrinsic parameters. As mentioned in the previous section, we use the axis angle representation for the rotation part $[r_x \quad r_y \quad r_z \quad r_{\theta}]$, and $[p_x \quad p_y\quad p_z]$ for the position part. 
\item Plane parameters. Each plane can be described by a unit vector $[{^B}n_{k,x}\quad {^B}n_{k,y}\quad {^B}n_{k,z}]$ normal to the plane and its perpendicular distance from the robot base's coordinate system origin ${^B}l_{k}$, so there are 12 parameters for 3 planes.
\end{enumerate}

In total, there are 43 parameters to be optimized by minimizing the objective function $f(\vb*{\Phi})$. To do that, the number of data points $3NM$ have to exceed the number of parameters. The optimization problem is then solved using a Levenberg-Marquardt nonlinear optimizer \cite{Newville2014}. The objective function $f(\vb*{\Phi})$ uses the geometric constraints that all data points from the \ac{lrf} have to fall on the respective plane. Zhuang et al. \cite{Zhuang1999} prove that a calibration process with such constraints is equivalent to the calibration of a robot using end-point measurement in unconstrained calibration. 

For the unit vector parameters ($[r_x \quad r_y \quad r_z ]$ and  $[{^B}n_{k,x} \quad {^B}n_{k,y} \quad {^B}n_{k,z}]$), the following constraints are added to the optimization solver:
\begin{equation}
\label{eq:10}
{r_z} = \sqrt{1 - {r_x}^2 - {r_y}^2}
\end{equation}
\begin{equation}
\label{eq:11}
{^B}n_{k,z} = \sqrt{1 - {{^B}n_{k,x}}^2 - {{^B}n_{k,y}}^2}
\end{equation}

Further analysis on the observability of the parameters will be presented in the next section.

\subsection{Identifiability of the calibration parameters}
\label{sec:third_step}

Depending on the chosen robot calibration poses and the robot's kinematic model, some of the calibration parameters might not be observable due to the linear dependency among the parameters. This is a critical problem in calibration, as it will result in some of the parameters assuming erratic values which gives us unstable calibration result. To prevent that, we have to first analyse which calibration parameters are identifiable and which are not. 

Following the approach in \cite{Joubair2015} and \cite{Hollerbach1996}, SVD is applied on the identification Jacobian matrix $\vb*{J}$. $\vb*{J}$ can be computed as follows. Let  $f_{kji}(\Phi)$ be the geometric constraint equation on the data point $i$ at the robot pose $j$ and on the plane k, 
\begin{equation}
\label{eq:12}
 f_{kji}(\vb*{\Phi}) =  {{^B}\vb*{n}_k}^T \cdot {^B}\vb*{p}_{ji} - {^B}l_k = 0 \; .
\end{equation}
Then $\vb*{J}$ can be computed by differentiating \eqref{eq:12} for all the data points $i = 1, \cdots, M$ at the robot poses $j = 1, \cdots, N$ and for all the planes $k=1,2,3$, then stack them together as a matrix,
\renewcommand\arraystretch{1.5}
\begin{equation}
\vb*{J} = \begin{bmatrix}
 \frac{\partial f_{111}(\vb*{\Phi})}{\partial\vb*{\Phi}} \quad
 \frac{\partial f_{112}(\vb*{\Phi})}{\partial\vb*{\Phi}} \quad
 \cdots  \quad
 \frac{\partial f_{3MN}(\vb*{\Phi})}{\partial\vb*{\Phi}} \quad
	\end{bmatrix} ^T \; .
\end{equation}

We can then apply SVD to the matrix $\vb*{J}$,
\begin{equation}
 \vb*{J} = \vb*{U}\vb*{\Sigma}\vb*{V}^T \; .
\end{equation}
Note that for this identification step, the parameters $[r_z \quad {^B}n_{1,z}\quad {^B}n_{2,z}\quad {^B}n_{3,z}]$ are excluded from the parameters vector $\vb*{\Phi}$, since those four parameters are obtained as linear combinations of other parameters (Equation \eqref{eq:10} and \eqref{eq:11}). That leaves us with 43-4 = 39 parameters in $\vb*{\Phi}$, which correlates to the 39 singular values in $\Sigma$. The number of zero singular values in $\Sigma$ is then equal to the number of unidentifiable parameters in the calibration procedure. For a given zero-value singular value $\sigma_r$, the r\textit{th} column vector of the matrix $\vb*{V}$ is the linear combination of the parameters $\vb*{\Phi}$ which cannot be identified independently.

\renewcommand{\arraystretch}{1.5}
\begin{table}[t]
\caption{DH Parameters of Denso VS060}
\label{tab:dh_params}
\centering
\begin{tabular}{c c c c c}
\toprule
i &  \textbf{$\alpha_i \;[^o]$} & \textbf{$a_i \;[mm]$} &  \textbf{$\theta_i \;[^o]$}  & \textbf{$d_i \;[mm]$}\\
\midrule
1 & 0.0 & 0.0 & $\theta_1$ & 345.0\\
2 & -90.0 & 0.0 & $\theta_2$ - 90.0 & 0.0\\
3 & 0.0 & 305.0 & $\theta_3$ + 90.0 & 0.0\\
4 & 90.0 & -10.0 & $\theta_4$ & 300.0\\
5 & -90.0 & 0.0 & $\theta_5$ & 0.0\\
6 & 90.0 & 0.0 & $\theta_6$ & 70.0\\
\bottomrule
\end{tabular}
\end{table}

In this paper, we use a Denso VS060 6-\ac{dof} industrial manipulator with its \textbf{DH} parameters presented in \tref{tab:dh_params}. The \ac{lrf}'s frame is defined such that the rotation part $[r_x \quad r_y \quad r_z \quad r_{\theta}] = [0 \quad 0 \quad 1 \quad \pi]$, and the position part $[p_x \quad p_y\quad p_z] = [-0.1275 \quad -0.033 \quad 0.1015]$. Applying the identifiability analysis to the system, we found that there are 7 sets of linearly dependent parameters out of the 39 parameters. 
\begin{enumerate}
\item The parameters $d_6$ (the translation along the z-axis of the 6\textit{th} link frame on the flange) and $p_z$ (the z coordinate of the \ac{lrf} frame) are linearly dependent. Physically this means that if we shift the origin of the 6\textit{th} link's frame  in its z direction by changing $d_6$, we can compensate by shifting the origin of the \ac{lrf} frame in the opposite direction by changing $p_z$.
\item The parameters $\theta_6$ and $r_\theta$ are linearly dependent. These correspond to the rotation of the 6\textit{th} link's frame and the rotation of the \ac{lrf}'s frame around the same z-axis. 
\item The parameters $d_2$ and $d_3$ are linearly dependent. These correspond to the shift in the z-axis direction of the 2\textit{nd} and 3\textit{rd} link's frames respectively, which are along the same direction. 
\item Lastly, we have four sets of linearly dependent parameters due to the linear combinations of the first link's DH parameters $[a_1, \alpha_1, \theta_1, d_1 ]$ and the three calibration planes' parameters. Physically, this relates to the fact that we can shift the robot's base frame freely by changing the value of $[a_1, \alpha_1, \theta_1, d_1]$, and the plane parameters will adjust according to the new base location. In other words, the base coordinate is not constrained (floating base). 
\end{enumerate}

For each set of the linearly dependent parameters, we can assign a fix value to one of the parameters. In this case, we fix the value of the parameters [$d_6, \theta_6, d_2, a_1, \alpha_1, \theta_1, d_1$] to their initial model's values. 

These results apply to most existing 6-\ac{dof} industrial robots whose structures are the same as that of our Denso robot.

\section{Simulation}
\label{sec:simulation}

We verify SCALAR through simulation of the calibration procedure. The simulation is conducted by using Robot Operating System and Gazebo where the robot model, the \ac{lrf}, and the three planes can be simulated.  
As shown in \fref{fig:robot_setup}, three perpendicular planes are located around the robot, and the robot is moved such that the \ac{lrf} ray intersects each plane. Simulated data from the \ac{lrf} can be obtained and Gaussian noise with zero mean and standar deviation $\sigma_{\rm{noise}}$ can be added to the data. The data is then used as input to the calibration procedure. 

After the calibration procedure, the robot is moved to 10,000 random poses to evaluate the accuracy of the calibrated parameters. Let ${^B}\vb*{T}_{L,j,true}$ and ${^B}\vb*{T}_{L,j,model}$ be the true and calibrated pose of the \ac{lrf}'s frame w.r.t. the robot base frame at the robot pose index $j$, respectively, then the error of the calibrated model can be computed as follows. 
\begin{equation}
\Delta \vb*{T}_j =  {{^B}\vb*{T}_{L,j,model}}^{-1} \cdot\; {^B}\vb*{T}_{L,j,true}
\end{equation}
Let $\delta t_{uv}$ be the element of $\Delta \vb*{T}_j$ with the subscript $u$ and $v$ refer to the row and column index, then \textbf{the position error} at the robot pose index $j$, $\delta p_j$, can be computed by
\begin{equation}
\delta p_j = \sqrt{\delta t_{14}^2 + \delta t_{24}^2 + \delta t_{34}^2} \quad .
\end{equation}
Let $\delta \vb*{R}_j$ be the rotation part of the homogeneous transformation matrix $\Delta \vb*{T}_j$. $\delta \vb*{R}_j$ can be represented by using an axis-angle notation, $[r_{j,1}\quad r_{j,2}\quad r_{j,3}\quad \delta \theta_j]$. We use $\delta\theta_j$ as \textbf{the orientation error} at the robot pose index $j$.  $\delta\theta_j$ can be seen as the amount of rotation necessary to rotate the calibrated pose to the true pose. The errors $\delta p_j$ and $\delta\theta_j$ are then averaged over the 10,000 random poses.

\begin{figure*}[t]
  \centering
  \subfloat[]{
  \includegraphics[height=38mm]{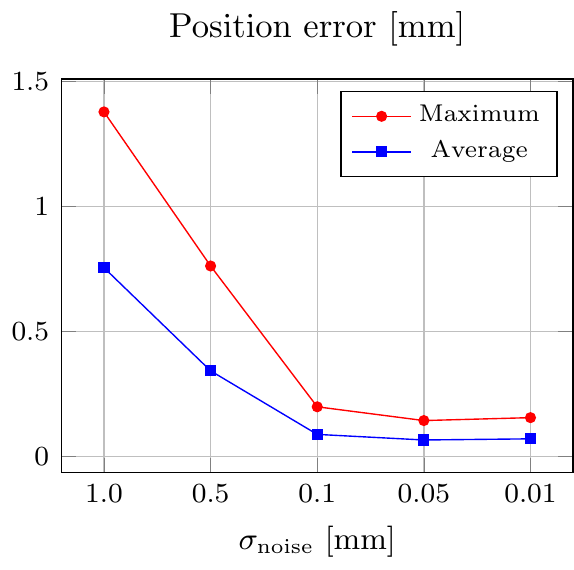}\;
  \includegraphics[height=38mm]{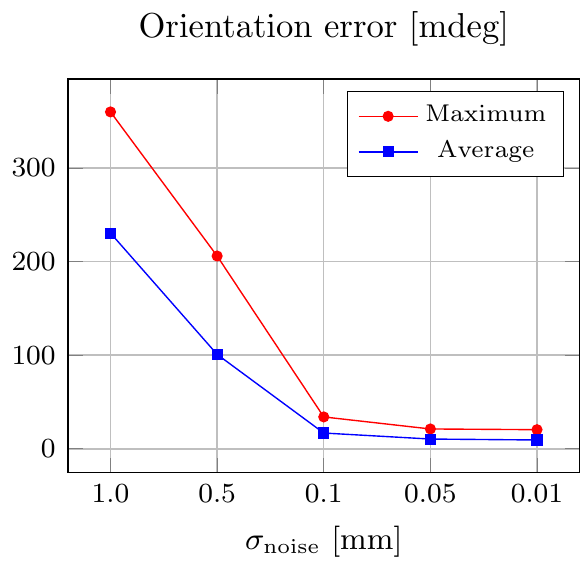}
  \label{fig:laser_noise}
  }\quad
  \subfloat[]{
  \includegraphics[height=38mm]{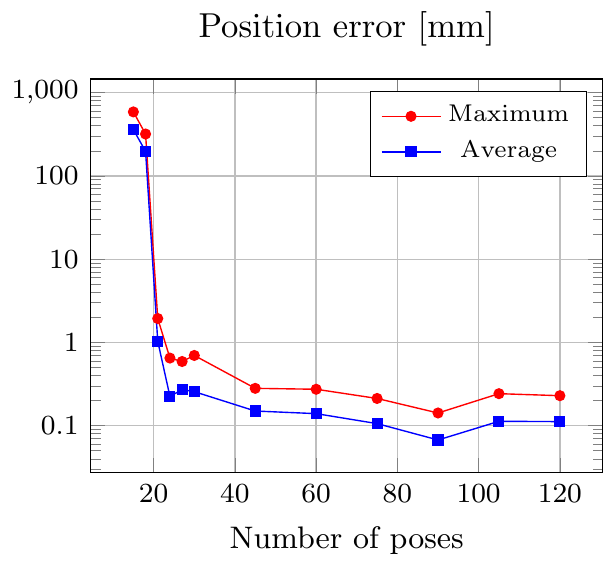}
  \includegraphics[height=38mm]{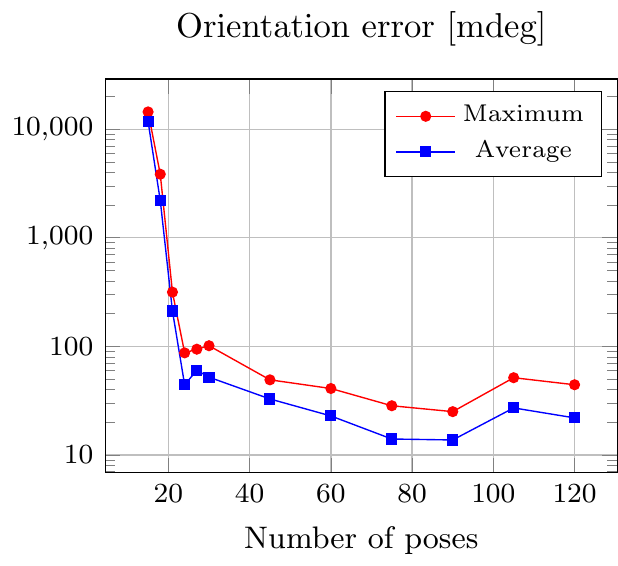}
  \label{fig:num_of_poses}
  }\\
  \centering
  \subfloat[]{
  \includegraphics[height=39mm]{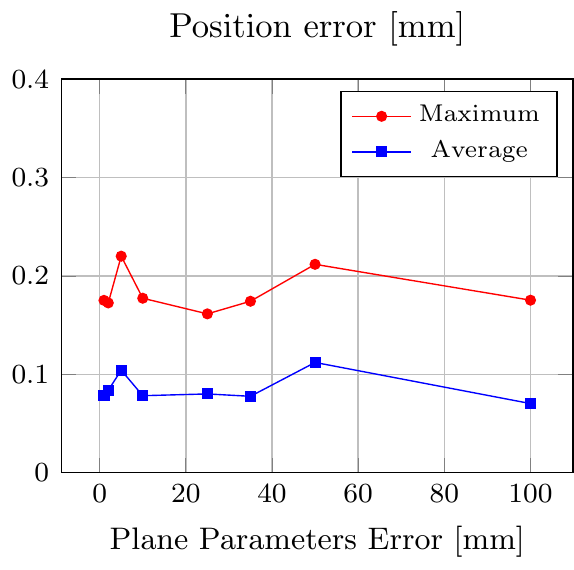}\;
  \includegraphics[height=39mm]{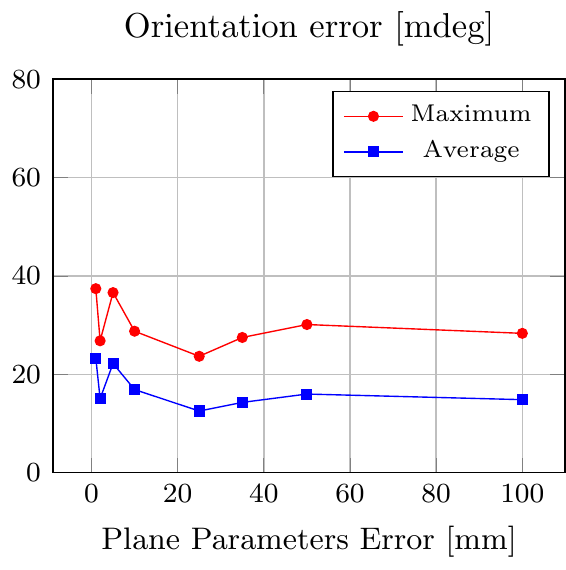}
  \label{fig:plane_params_linear}
  }\quad\quad
  \subfloat[]{
  \includegraphics[height=39mm]{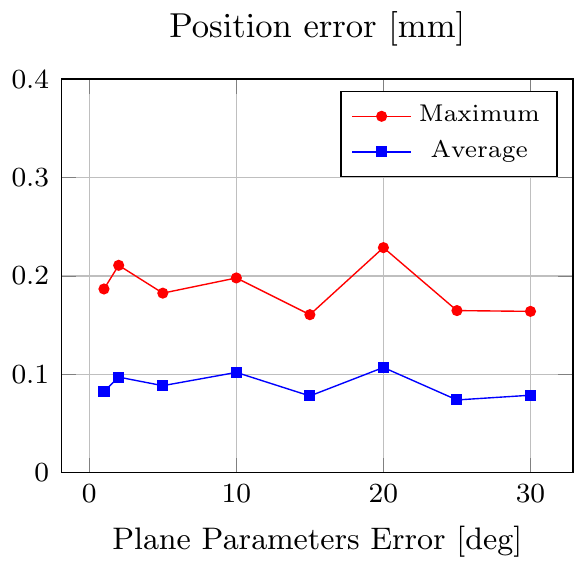}\quad
  \includegraphics[height=39mm]{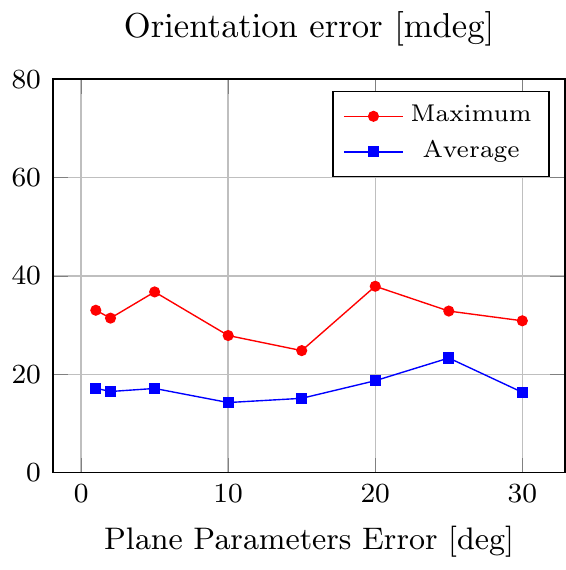}
  \label{fig:plane_params_angular}
  }\quad  
   \caption{Effect of a) the measurement noise, b) the number of poses, c) planes' position estimate error, and d) plane's orientation estimate error towards the position and orientation error after calibration}
\end{figure*}

The simulated robot's model is considered as having the true kinematics and extrinsic parameters, and an initial model is generated by introducing random Gaussian errors to the true parameters within the range of $\pm$ 2mm and $\pm$ 1$^o$ for the linear and angular parameters, respectively. Note that the initial model's errors are intentionally set to be large to illustrate the robustness of the calibration method. The average position and orientation errors of the initial model as compared to the true model are 14.6mm and 4.05$^o$, while the maximum errors are 103.9mm and 5.95$^o$. We run the calibration procedure to improve the initial model with $3N$ = 120, $M$ = 100 and $\sigma_{\rm{noise}}$ = 0.1mm, and the resulting calibrated model has the average position and orientation errors reduced to 0.09mm and 0.02$^o$, while the maximum errors are reduced to 0.19mm and 0.035$^o$.

Next, we evaluate the effect of the measurement noise, the number of poses ($N$) and points ($M$), and the plane parameters' initial estimate on the calibration errors.

\subsection{Effect of the measurement noise}
\label{sec:meas_accuracy}
The accuracy of the calibration procedure greatly depends on the accuracy of the measurement system, which is affected by the noise on the data. In this section, Gaussian noises with zero means and varying standard deviations $\sigma_{\rm{noise}}$ are added to the measurement data in the simulation, and its effect on the calibration errors is shown in \fref{fig:laser_noise}. As $\sigma_{\rm{noise}}$ decreases, the calibration errors decrease. At $\sigma_{\rm{noise}}$ = 0.1mm, the average position and orientation errors are around 0.1mm and 0.02$^o$, respectively. For the subsequent sections, $\sigma_{\rm{noise}}$ is set at 0.1mm. 

\subsection{Effect of the number of calibration poses and the number of points}
\label{sec:calib_poses}
For each plane, the robot is moved to $N$ poses, so in total there are $3N$ calibration poses. At each pose, we select $M$ data points from the \ac{lrf} data. In this section we evaluate the effect of $3N$ and $M$ to the calibration errors. In \fref{fig:num_of_poses}, it can be seen that as the number of poses $3N$ increases, the error decreases until around $3N$=60, beyond which it does not change significantly. It can be concluded that 60 robot poses are sufficient to calibrate the robot model accurately. We also conduct similar analysis on $M$ (the data is not presented in this paper) and found that $M$ = 20 is sufficient for the calibration.

\subsection{Effect of the plane parameters' initial estimate}
\label{sec:plane_params}
One of the advantages of SCALAR is that the plane parameters do not need to be precisely known. Here we vary the plane parameters' estimate to demonstrate the robustness of our method. The initial estimates of the positions and the normals of the planes are disturbed by up to 100mm and 30$^o$, as shown in \fref{fig:plane_params_linear} and \fref{fig:plane_params_angular}. From the figures, it can be seen that the calibration position and orientation errors are not affected by the errors in the plane parameters' initial estimate. In fact, after calibration, the plane parameters in the calibrated model approach the true parameters within 0.1mm and 0.01$^o$.

\section{Conclusions}
\label{sec:conclusions}

In this paper, we have proposed SCALAR, a method to simultaneously calibrate a 6-\ac{dof} robot's kinematic parameters and a 2D \ac{lrf}'s extrinsic parameters using only three flat planes, arranged perpendicularly towards each other around the robot. SCALAR is easier to implement than the previous methods in the literature as it does not require other expensive measurement systems or tedious setup. Through simulations, we have also verified that the method can reduce the average errors in position and orientation from (14.6mm, 4.05$^o$) to (0.09mm, 0.02$^o$), respectively. This is very useful for many industrial robotics applications that require great accuracy. 

The next step after this is to implement SCALAR on the real system. Some challenges that may appear on the real system are the backlash of the robot's transmission system, the elasticity of the joints, the roughness of the calibration planes and the noises of the \ac{lrf} data which will reduce the calibration accuracy. Moreover, evaluating the calibration errors in the real system is not as easy as in simulation. We will present the calibration result on the real system in our future work.

\section*{Acknowledgment}
This work was supported in part by NTUitive Gap Fund NGF-2016-01-028
and SMART Innovation Grant NG000074-ENG.

\IEEEtriggeratref{7}
\bibliographystyle{IEEEtran}
\bibliography{IEEEabrv,references}

\end{document}